\documentclass[10pt,twocolumn,letterpaper]{article}

\usepackage{cvpr}
\usepackage{times}
\usepackage{epsfig}
\usepackage{graphicx}
\usepackage{amsmath}
\usepackage{amssymb}
\usepackage[numbers,sort&compress]{natbib}

\usepackage[breaklinks=true,bookmarks=false]{hyperref}

\cvprfinalcopy 

\def\cvprPaperID{****} 

\newif\ifsqueeze
\squeezetrue  
\ifsqueeze

\else

\fi

\begin{document}

\title{The iWildCam 2019 Challenge Dataset}

\author{Sara Beery$^{*}$, Dan Morris$^{+}$, Pietro Perona$^{*}$\\
California Institute of Technology\\
1200 E California Blvd., Pasadena, CA 91125\\
}

\maketitle

\begin{abstract}
Camera Traps (or Wild Cams) enable the automatic collection of large quantities of image data. Biologists all over the world use camera traps to monitor biodiversity and population density of animal species. The computer vision community has been making strides towards automating the species classification challenge in camera traps, but as we try to expand the scope of these models from specific regions where we have collected training data to different areas we are faced with an interesting problem: how do you classify a species in a new region that you may not have seen in previous training data? 

In order to tackle this problem, we have prepared a dataset and challenge where the training data and test data are from different regions, namely The American Southwest and the American Northwest. We use the Caltech Camera Traps dataset, collected from the American Southwest, as training data. We add a new dataset from the American Northwest, curated from data provided by the Idaho Department of Fish and Game (IDFG), as our test dataset. The test data has some class overlap with the training data, some species are found in both datasets, but there are both species seen during training that are not seen during test and vice versa.  To help fill the gaps in the training species, we allow competitors to utilize transfer learning from two alternate domains: human-curated images from iNaturalist and synthetic images from Microsoft's TrapCam-AirSim simulation environment.
\end{abstract}
\section{Introduction}

\begin{figure}[h!]
    \centering
    \includegraphics[width=7.2cm]{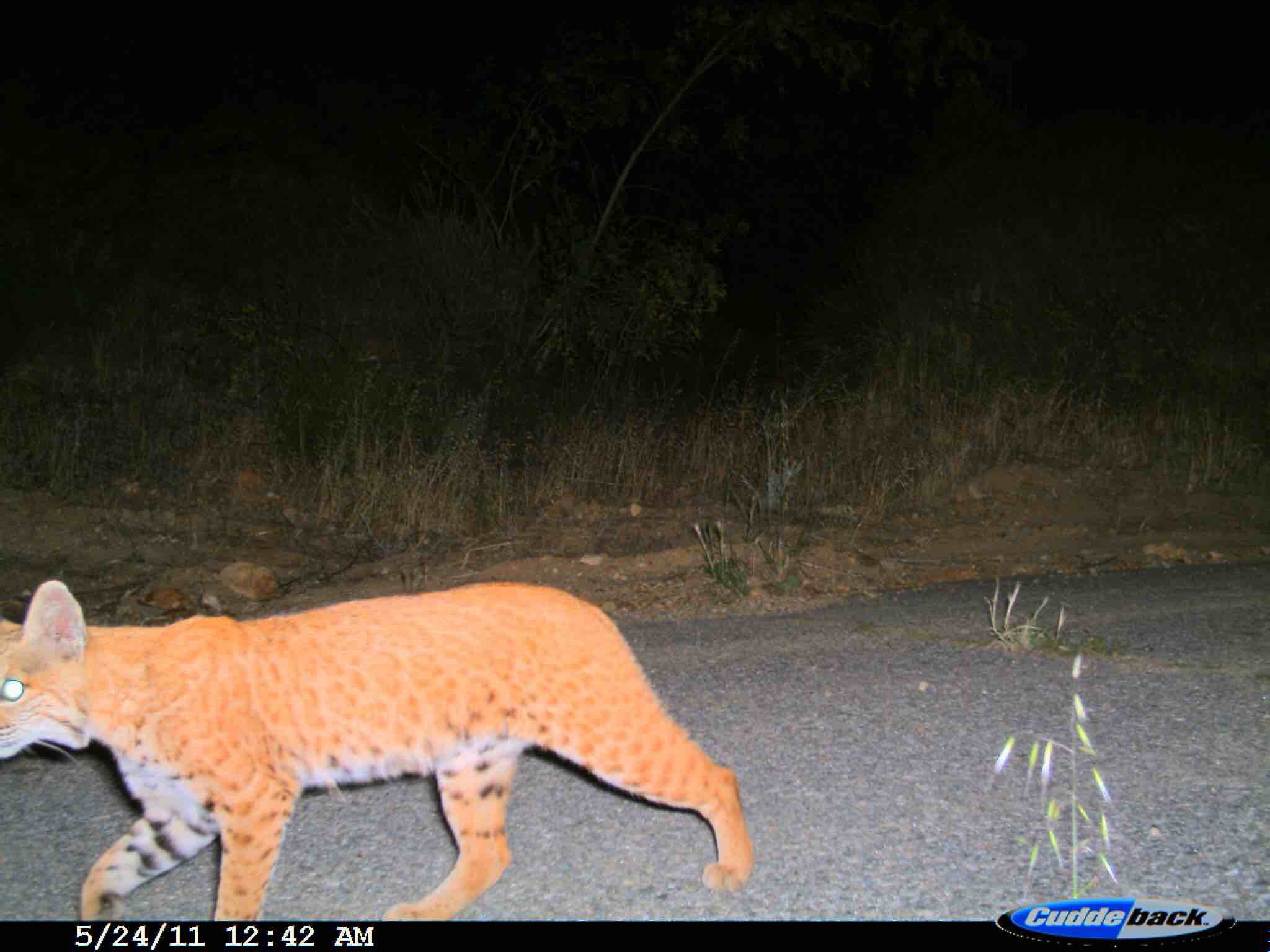}\\
    \vspace{5pt}
    \includegraphics[width=7.2cm]{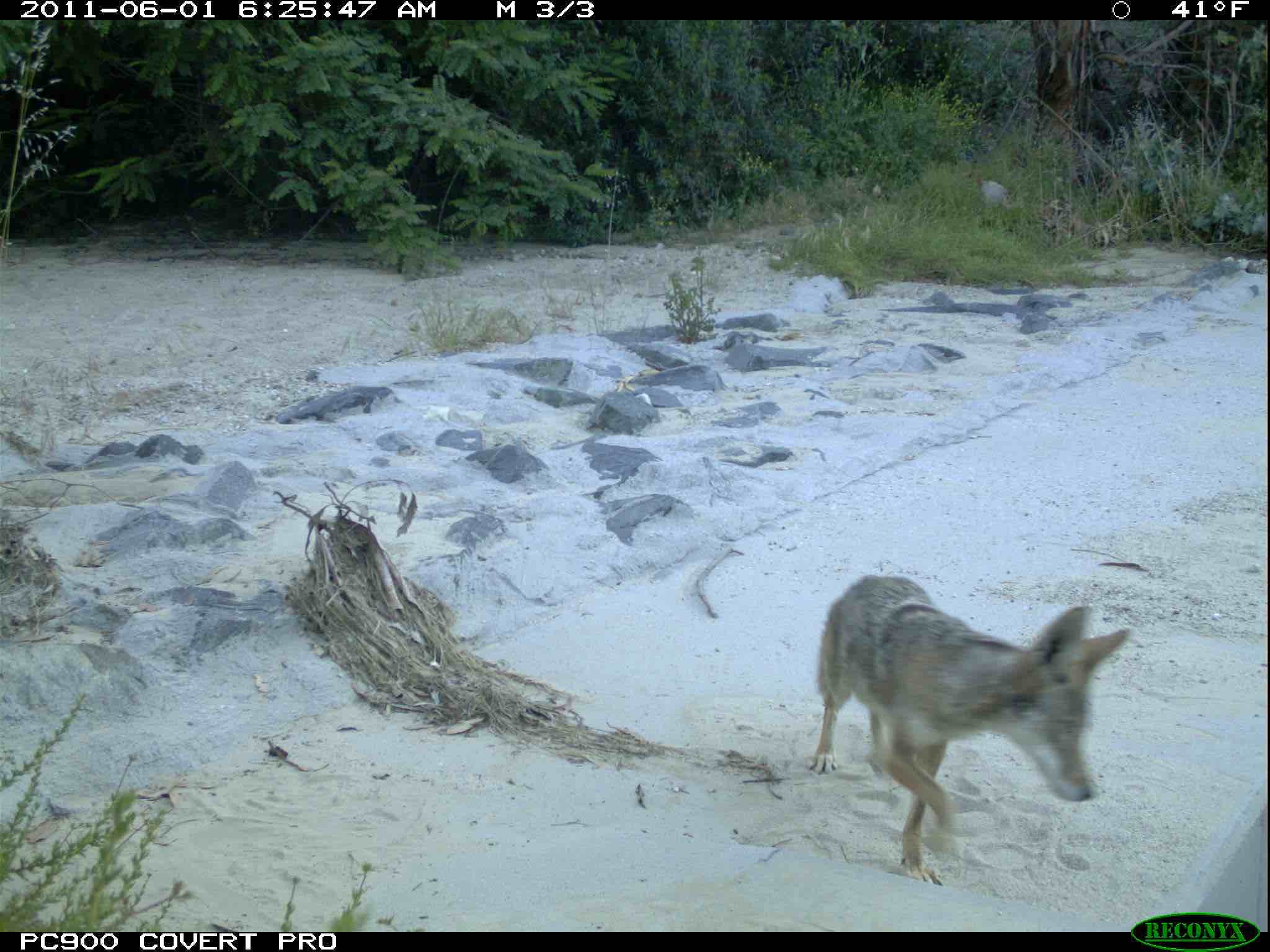}\\
    \vspace{5pt}
    \includegraphics[width=7.2cm]{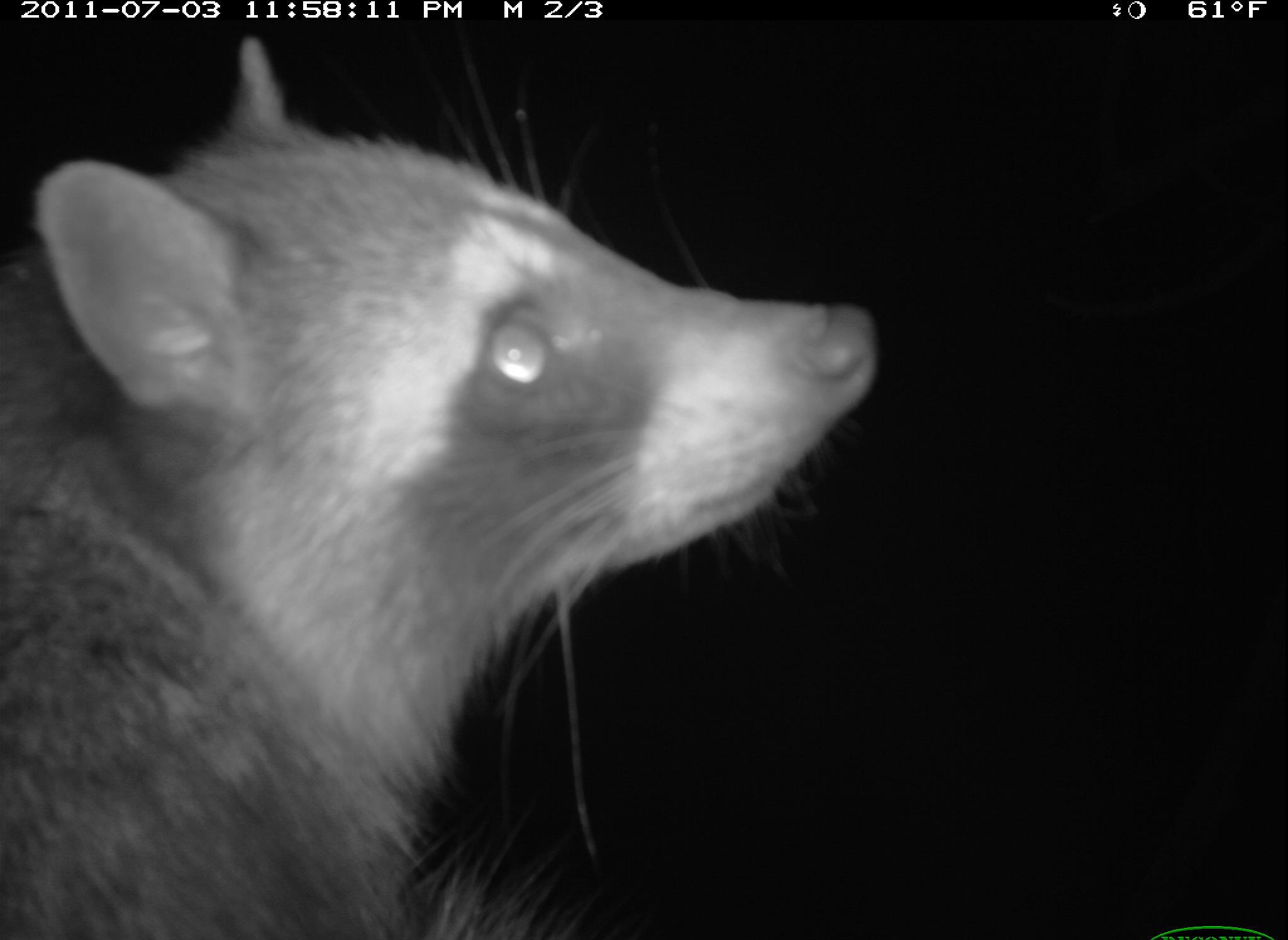}
    \caption{Examples of camera trap data.}
    \label{fig:camera trap data}
\end{figure}

Monitoring biodiversity quantitatively can help us understand the connections between species decline and pollution, exploitation, urbanization, global warming, and conservation policy. Researchers study the effect of these factors on wild animal populations by monitoring changes in species diversity, population density, and behavioral patterns using \textit{camera traps}: heat- or motion-activated cameras placed in the wild (See Fig.~\ref{fig:camera trap data} for examples). 

\begin{figure}
\begin{minipage}[b]{.3\linewidth}
  \centering
  \centerline{\includegraphics[width=3cm]{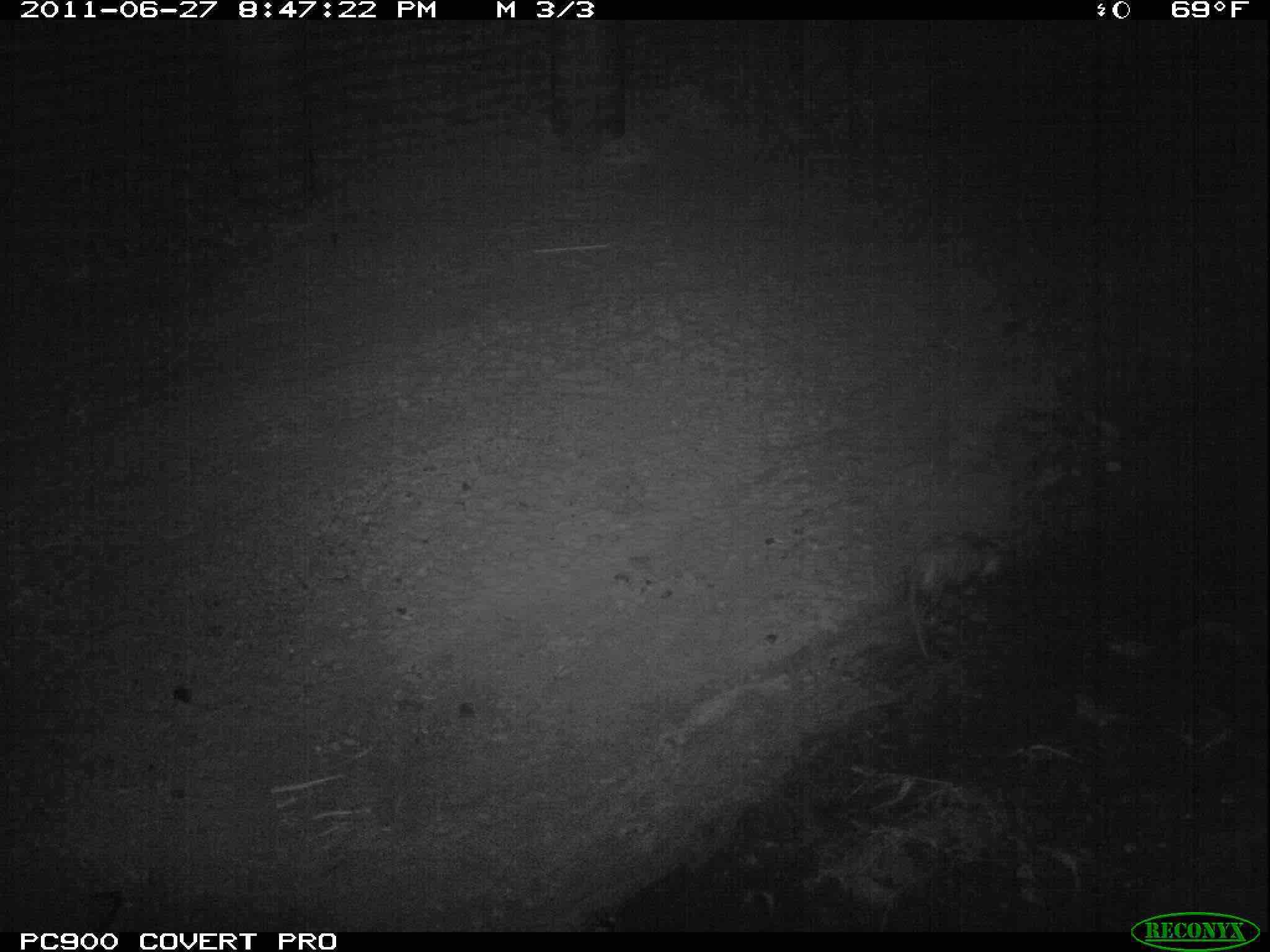}}
  \vspace{.05cm}
  \centerline{(1) Illumination}\medskip
\end{minipage}
\hfill
\begin{minipage}[b]{0.3\linewidth}
  \centering
  \centerline{\includegraphics[width=3cm]{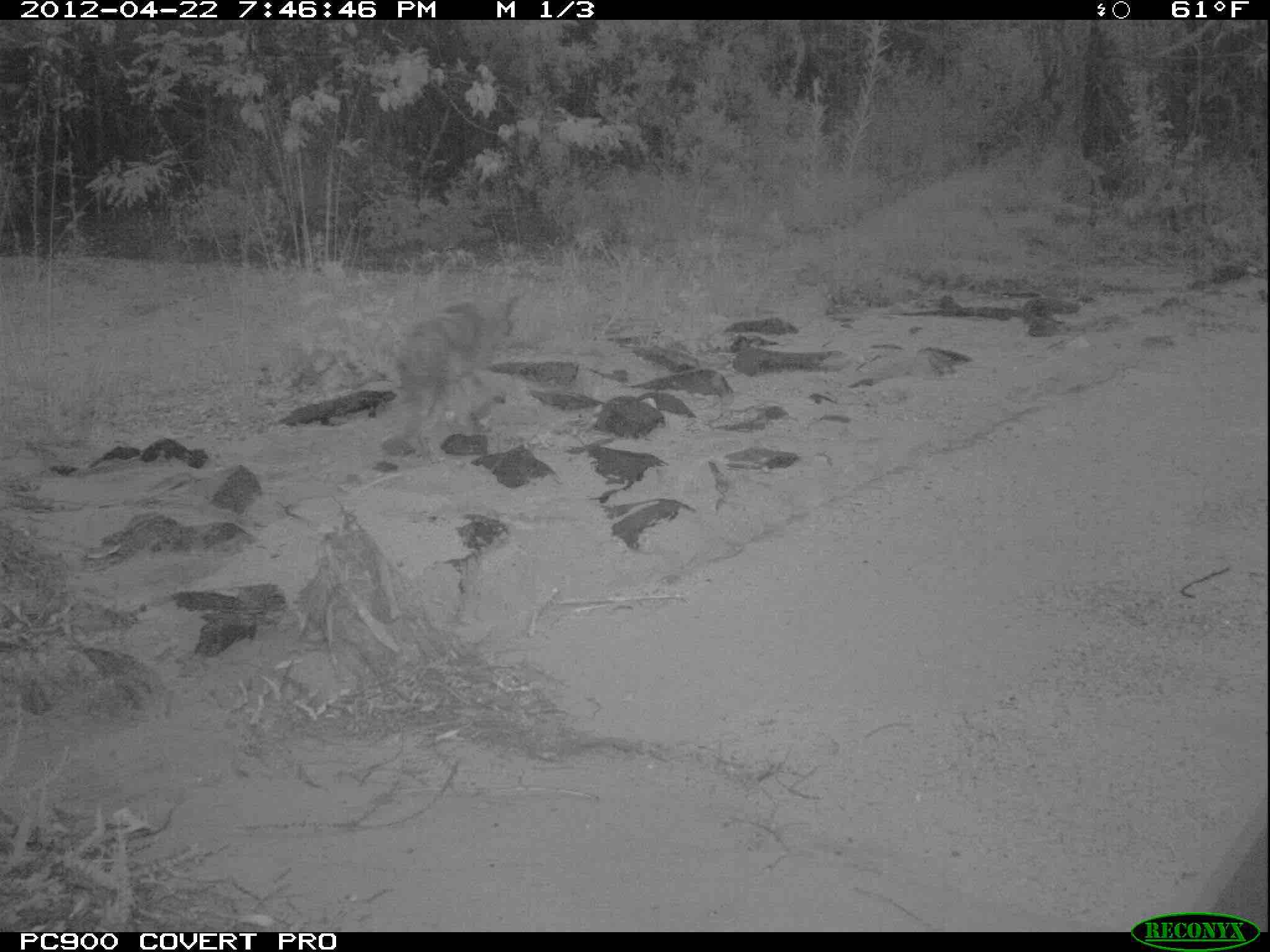}}
  \vspace{.05cm}
  \centerline{(2) Blur}\medskip
\end{minipage}
\hfill
\begin{minipage}[b]{.3\linewidth}
  \centering
  \centerline{\includegraphics[width=3cm]{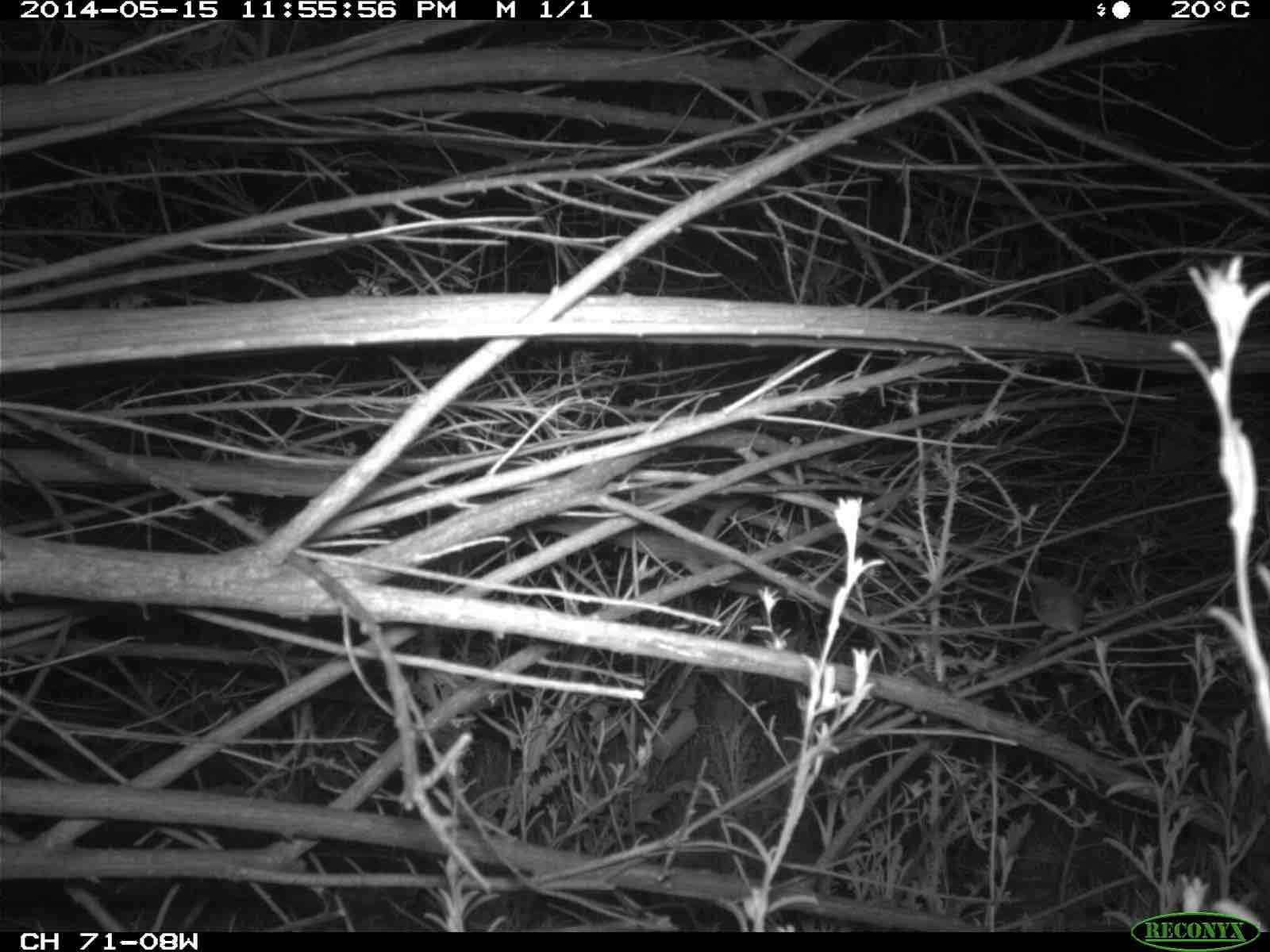}}
  \vspace{.05cm}
  \centerline{(3) ROI Size}\medskip
 \end{minipage}
\begin{minipage}[b]{0.3\linewidth}
  \centering
  \centerline{\includegraphics[width=3cm]{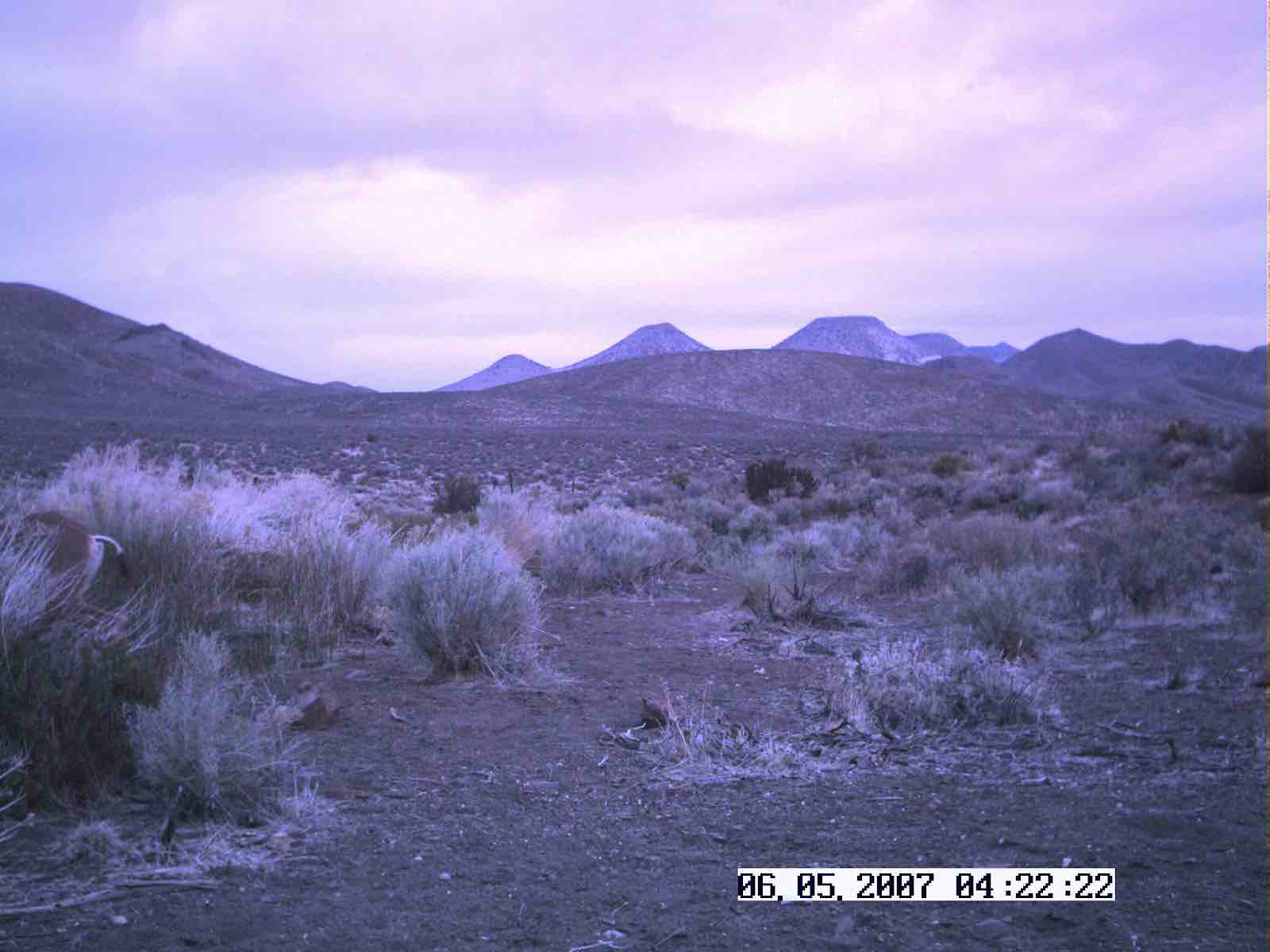}}
  \centerline{(4) Occlusion}\medskip
\end{minipage}
\hfill
\begin{minipage}[b]{.3\linewidth}
  \centering
  \centerline{\includegraphics[width=3cm]{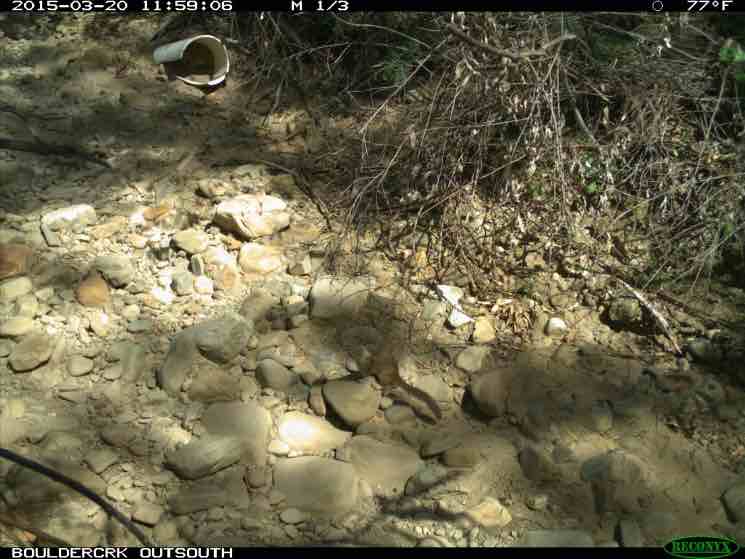}}
  \centerline{(5) Camouflage}\medskip
\end{minipage}
\hfill
\begin{minipage}[b]{0.3\linewidth}
  \centering
  \centerline{\includegraphics[width=3cm]{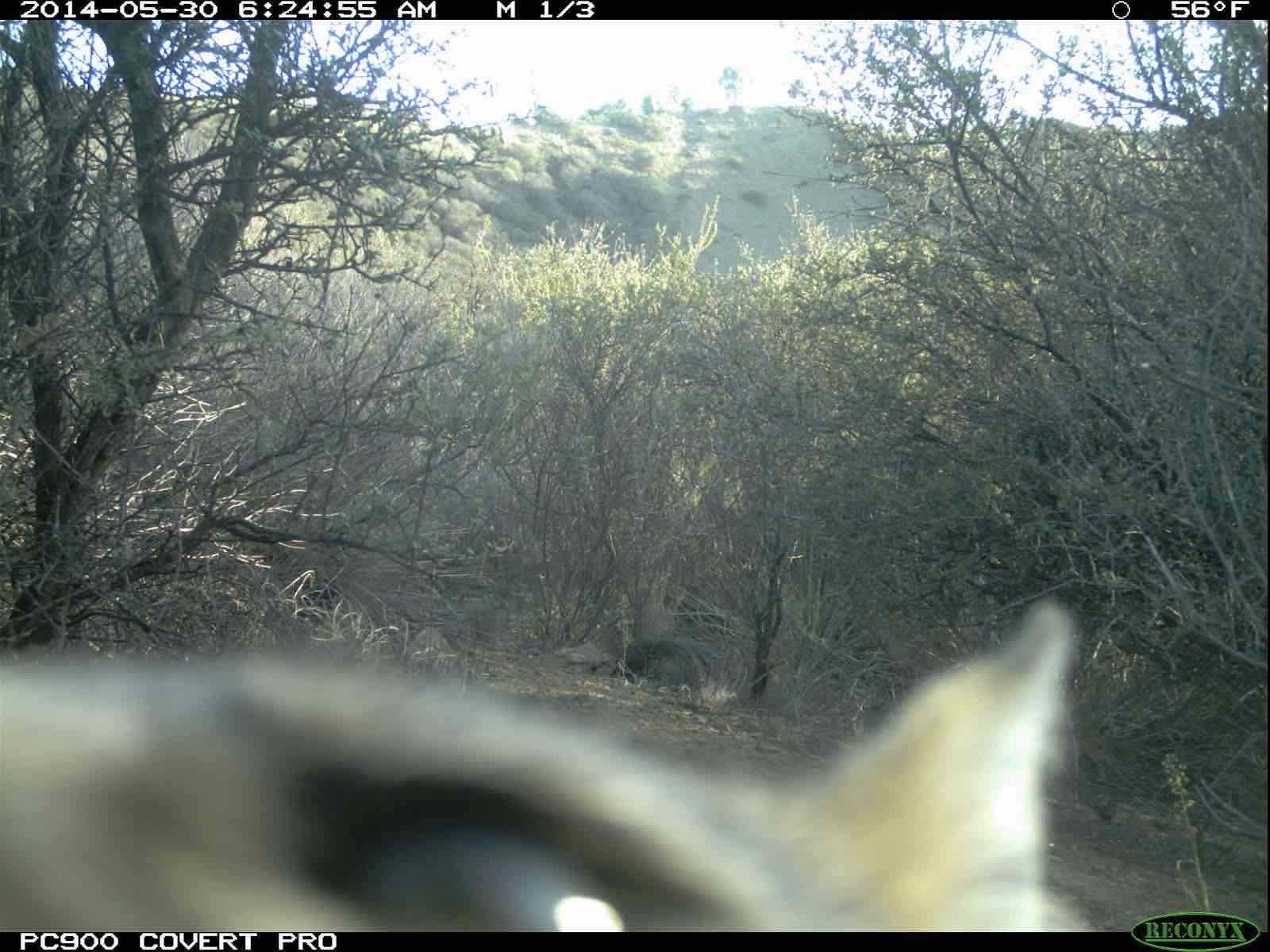}}
  \centerline{(6) Perspective}\medskip
\end{minipage}
\caption{\textbf{Common data challenges}: (1) {\bf Illumination}: Animals are not always salient. (2) {\bf Motion blur}: common with poor illumination at night. (3) {\bf Size of the region of interest} (ROI): Animals can be small or far from the camera. (4) {\bf Occlusion}: e.g. by bushes or rocks. (5) {\bf Camouflage}: decreases saliency in animals' natural habitat. (6) {\bf Perspective}: Animals can be close to the camera, resulting in partial views of the body.}
\label{fig:challenging_ims}
\end{figure}

\clearpage
At present, camera trap images are annotated by hand, severely limiting data scale and research productivity. Annotation of camera trap photos is time consuming and challenging. Because the images are taken automatically based on a triggered sensor, there is no guarantee that the animal will be centered, focused, well-lit, or an appropriate scale (they can be either very close or very far from the camera, each causing its own problems). See Fig. \ref{fig:challenging_ims} for examples of these challenges. Further, up to 70\% of the photos at any given location are triggered by something other than an animal, such as wind in the trees, a passing car, or a hiker. 

Automating camera trap labeling is not a new challenge for the computer vision community \cite{ren2013ensemble,yu2013automated,wilber2013animal,chen2014deep,lin2014foreground,swanson2015snapshot,zhang2015coupled,zhang2016animal,miguel2016finding,giraldo2017camera,yousif2017fast,villa2017towards,norouzzadeh2017automatically, beery2018recognition, beery2019synthetic}. 
However, most of the proposed solutions have used the same camera locations for both training and testing the performance of an automated system. If we wish to build systems that are trained once to detect and classify animals, and then deployed to new locations without further training, we must measure the ability of machine learning and computer vision to {\em generalize} to new environments. Both the 2018 \cite{beery2019iwildcam} and 2019 iWildCam challenges focus on generalization, and encourage computer vision researchers to tackle generalization in creative, novel ways.

\section{The iWildCam 2019 Dataset}
The data for the 2019 challenge is curated from the Caltech Camera Traps (CCT) which was also used for the iWildCam 2018 Challenge~\cite{beery2019iwildcam}, a new camera trap dataset from Idaho (IDFG), and two alternate data domains: iNaturalist and Microsoft TrapCam-AirSim.

\subsection{Caltech Camera Traps}
All images in this dataset, which was used for the iWildCam 2018 Challenge, come from the American Southwest. By limiting the geographic region, the flora and fauna seen across the locations remain consistent. Examples of data from different locations can be seen in Fig.~\ref{fig:camtrap_ims}. 
This dataset consists of $292,732$ images across $143$ locations, each labeled with an animal class, or as empty. The classes represented are bobcat, opossum, coyote, raccoon, dog, cat, squirrel, rabbit, skunk, rodent, deer, fox, mountain lion, empty. We do not filter the stream of images collected by the traps, rather this is the same data that a human biologist currently sifts through. Therefore the data is unbalanced in the number of images per location, distribution of species per location, and distribution of species overall (see Fig.~\ref{fig:annotPerLoc}). The class of each image was provided by expert biologists from the NPS and USGS. Due to different annotation styles and challenging images, we approximate that the dataset contains up to 5\% annotation error.

\subsection{IDFG}
\begin{figure}
\centering
\includegraphics[height=5cm]{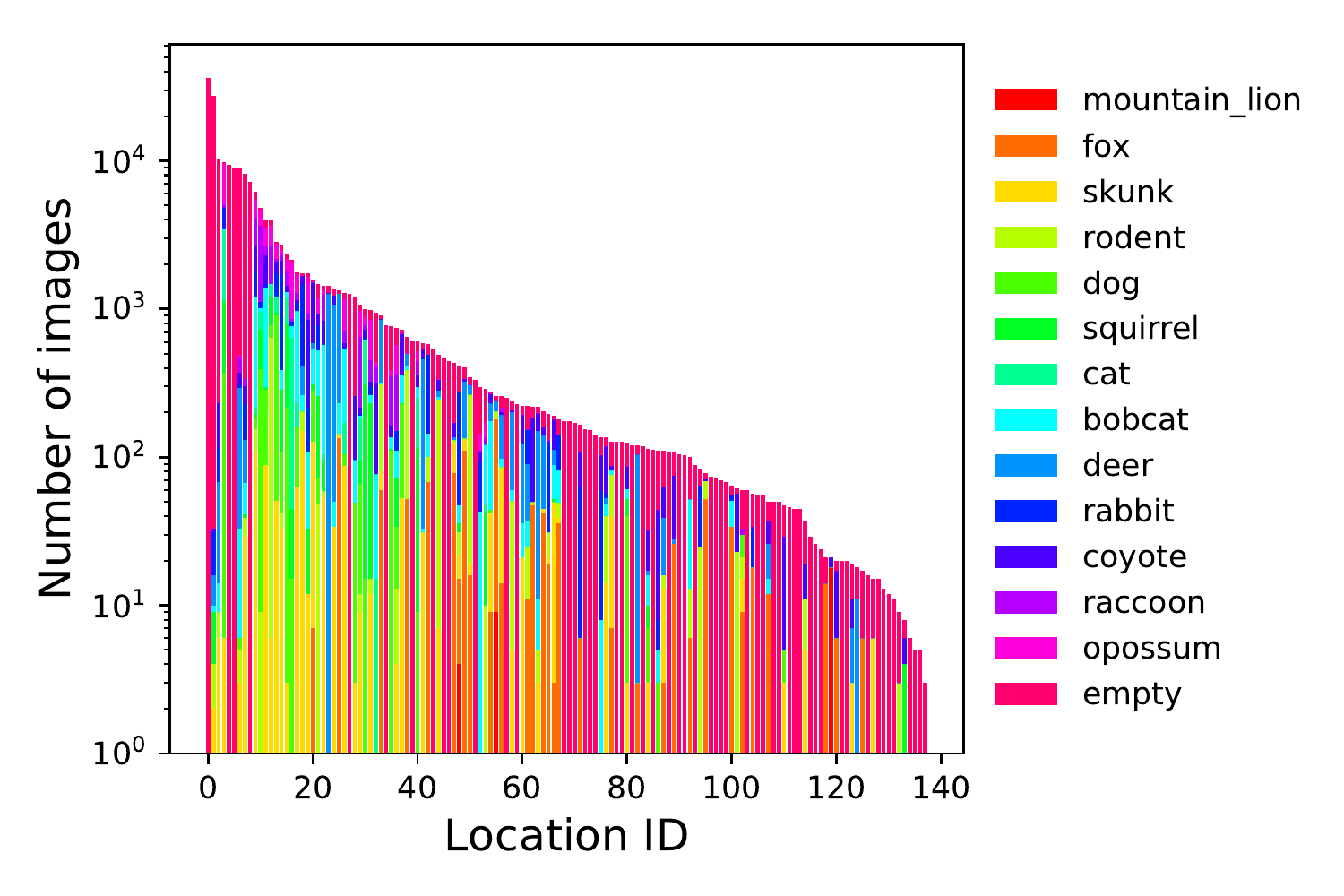} 
\includegraphics[height=5cm]{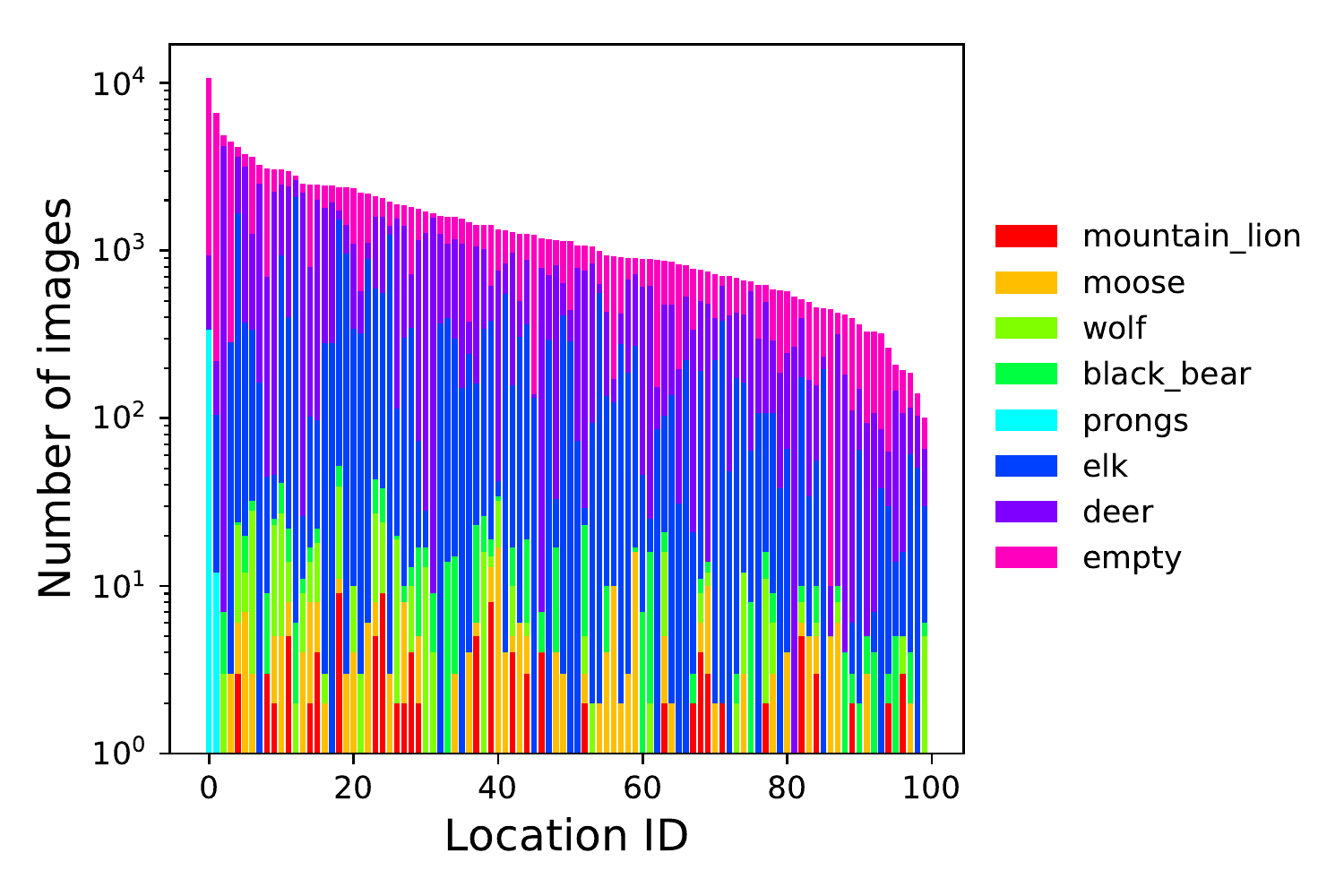} 
\caption{\textbf{Number of annotations for each location}. (Top) CCT locations, containing 14 classes. (Bottom) IDFG locations, containing images of 8 classes. The distribution of images per location is long-tailed, and each location has a different and peculiar class distribution.}
\label{fig:annotPerLoc}
\end{figure}

The Idaho Department of Fish and Game provided labeled data from Idaho to use as an unseen test set, which we call IDFG.  The test set contains 153,730 images from 100 locations in Idaho.  It covers the classes mountain lion, moose, wolf, black bear, pronghorn, elk, deer, and empty. See Fig.~\ref{fig:annotPerLoc} for the distribution of classes and images across locations. Similarly to CCT, we do not filter the images so the data is innately unbalanced.  

\subsection{Additional Data Domains}

\subsubsection{iNaturalist}
iNaturalist is a website where citizen scientists can post photos of plants and animals and work together to correctly ID the photos, an example of an iNaturalist image can be seen in Fig.~\ref{fig:domain}. We allow the use of iNaturalist data from both the 2017 and 2018 iNaturalist competition datasets \cite{van2018inaturalist}. For ease of entry, we did the work to map our classes into the iNaturalist taxonomy. We also determined which mammals might be seen in Idaho using the iNaturalist API: bobcat, opossum, coyote, raccoon, dog, cat, squirrel, rabbit, skunk, rodent, deer, fox, mountain lion, moose, small mammal, elk, pronghorn, bighorn sheep, black bear, wolf, bison, and mountain goat. We curated an iNat-Idaho dataset that contains all iNat classes that might occur in Idaho, mapped into our class set in order to make adapting iNaturalist data for this challenge as simple as possible.

\subsubsection{Microsoft TrapCam-AirSim}
\begin{figure}
\centering
\includegraphics[height=3cm]{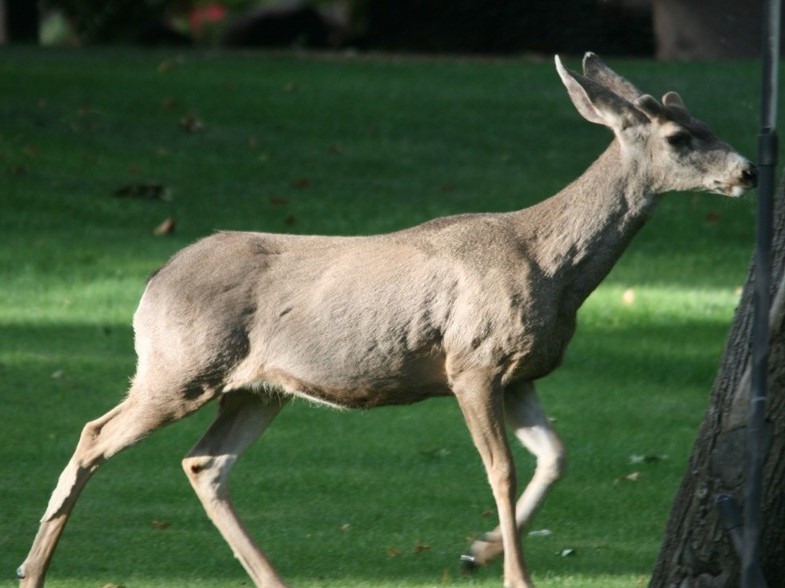} 
\includegraphics[height=3cm]{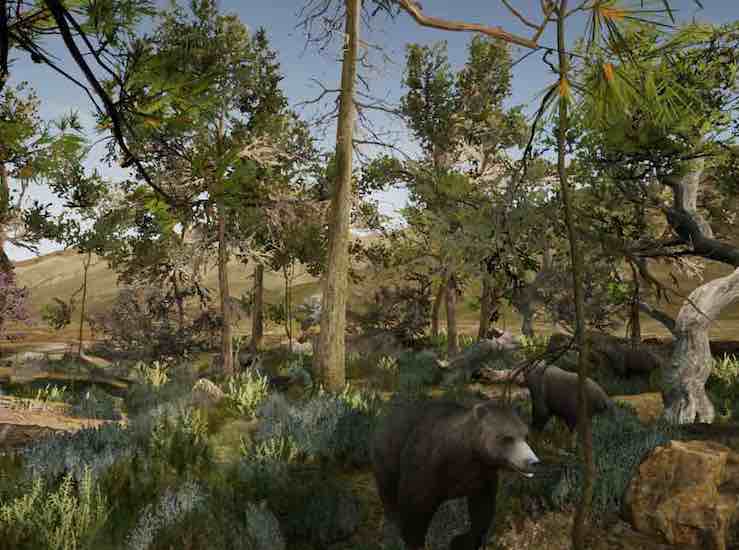} 
\caption{\textbf{Altenate domain examples.} (Left) iNaturalist, (Right) TrapCam-AirSim }
\label{fig:domain}
\end{figure}

This synthetic data generator utilizes a modular natural environment within Microsoft AirSim~\cite{shah2018airsim, beery2019synthetic} that can be randomly populated with flora and fauna.  The distribution and types of animals, trees, bushes, rocks, and logs can be varied and randomly seeded to create images from a diverse set of classes and landscapes, from an open plain to a dense forest. An example of a TrapCam-AirSim image containing a bison can be seen in Fig.~\ref{fig:domain}.

\section{The iWildCam Challenge 2019}
The iWildCam Challenge 2019 was conducted through Kaggle as part of FGVC6 at CVPR19. We used macro-average F1 score as our competition metric, to slightly emphasize recall over precision and to encourage more emphasis on rare classes, as opposed to rewarding high performance on common classes proportionally to their unbalanced level of occurrence.

\subsection{Data Split and Baseline}
We do not explicitly define a validation set for this challenge, instead letting competitors create their own validation set from the CCT training set and the two external data domains, iNat and TrapCam-AirSim.  We use the IDFG data as our test set. Unsupervised annotation of the test set, using the provided detector or any clustering methods, is allowed. Explicit annotation of the test set is not.  

We trained a simple whole-image classification baseline using the Inception-Resnet-V2 architecture, pretrained on ImageNet and trained simultaneously on the CCT and iNat-Idaho datasets with no class rebalancing or weighting, with an initial learning rate of 0.0045, rmsprop with a momentum of 0.9, and a square input resolution of 299. We employed random cropping (containing most of the region), horizontal flipping, and random color distortion as data augmentation. This baseline achieved 0.125 macro-averaged F1 score and accuracy of 27.6\% on the IDFG test set.

\subsection{Camera Trap Animal Detection Model}
We also provide a general animal detection model which competitors are free to use as they see fit. The model is a tensorflow Faster-RCNN model with Inception-Resnet-v2 backbone and atrous convolution. Sample code for running the detector over a folder of images can be found at https://github.com/Microsoft/CameraTraps. We have run the detector over each dataset, and provide the top 100 boxes and associated confidences for each image.
\section{Conclusions}
Camera traps provide a unique experimental setup that allow us to explore the generalization of models while controlling for many nuisance factors. This dataset provides a test bed for studying generalization to not only new geographic regions, but also new species not seen during training. This dataset is the first designed to explicitly study generalization to a new region with a non-identical class set. The difficulty of the associated challenge demonstrates the innate inability for our current state-of-the-art methods to handle this type of generalization. 

In subsequent years, we plan to extend the iWildCam Challenges by adding new regions and species worldwide. We hope to use the knowledge we gain throughout these challenges to facilitate the development of models or systems of models that can accurately provide real-time species ID in camera trap images at a global scale. Any forward progress made will have a direct impact on the scalability of biodiversity research geographically, temporally, and taxonomically.

{\small
\bibliographystyle{ieee}
\bibliography{main}
}

\end{document}